\documentclass[10pt,twocolumn,a4paper]{esaAI}
\usepackage{subcaption}
\usepackage{draftwatermark}
\SetWatermarkText{Accepted Pre-print}
\SetWatermarkScale{0.5}
\title{On-orbit Servicing for Spacecraft Collision Avoidance With Autonomous Decision Making}

\def\authorEmail{adam.abdin@centralesupelec.fr}

\author
[]{Susmitha Patnala}
\author[]{Adam Abdin \thanks{Corresponding author. E-Mail: \authorEmail}}

\affil[]{Université Paris-Saclay, CentraleSupélec, Industrial Engineering Research Department, Gif-sur-Yvette, France}

\begin{document}

\makeCustomtitle

\begin{abstract}
This study develops an AI-based implementation of autonomous On-Orbit Servicing (OOS) mission to assist with spacecraft collision avoidance maneuvers (CAMs). We propose an autonomous `servicer' trained with Reinforcement Learning (RL) to autonomously detect potential collisions between a target satellite and space debris, rendezvous and dock with endangered satellites, and execute optimal CAM. The RL model integrates collision risk estimates, satellite specifications, and debris data to generate an optimal maneuver matrix for OOS rendezvous and collision prevention. We employ the Cross-Entropy algorithm to find optimal decision policies efficiently. Initial results demonstrate the feasibility of autonomous robotic OOS for collision avoidance services, focusing on one servicer spacecraft to one endangered satellite scenario. However, merging spacecraft rendezvous and optimal CAM presents significant complexities. We discuss design challenges and critical parameters for the successful implementation of the framework presented through a case study.
\end{abstract}
\section{Introduction}\label{sec:intro}
Space exploration and satellite deployment have become increasingly common in recent decades. As a result, the space environment around Earth becomes increasingly congested with both active spacecraft and space debris, with over 36,500 pieces of space debris larger than 10 centimeters in diameter \cite{esa2023}. This debris poses significant risks to manned and unmanned spacecraft, making it a pressing concern for spacecraft operators around the world.  \\
To manage spacecraft collision challenges, space operators and researchers have developed tools for conjunction assessment and mitigation strategies \cite{kim2018development}. For example, NASA's Robotic Conjunction Assessment Risk Analysis framework has been identifying and reacting to predicted close approaches for many spacecraft \cite{newman2014evolution}. The International Space Station routinely performs collision risk assessments and avoidance maneuvers \cite{nasadebris3}. Strategies to assess Collision Avoidance Maneuvers (CAMs) for Cryosat-2 were started after a near-miss incident with ERS-2 \cite{symonds2014operational}. \\
With the significant growth of satellite constellations, autonomous collision avoidance systems are becoming imperative. Efforts in satellite collision avoidance have evolved from early orbit propagation algorithms, such as RBF-collocation, to sophisticated space-based situational awareness technologies and collision avoidance services, such as ESA's CORAM and Occam \cite{elgohary2015rbf, wang2022research, merz2021esa, cobo2014coram, hernando2016occam}.  Recent advances in AI and machine learning, particularly reinforcement learning (RL), have shown promise in space traffic management \cite{sanchez2020ai, gremyachikh2019space,sanchez2019ai} and autonomous spacecraft docking \cite{willis2016reinforcement, wang2018autonomous, anderlini2019docking, oestreich2020autonomous, oestreich2021autonomous}. However, while several research works investigated the potential of On-Orbit Service (OSS) for purposes such as satellite refueling, repairs, and active debris capture \cite{easdown2020mission,nasa2010orbit,grover2008development}, to our knowledge, no study has analyzed the feasibility of autonomous robotic OOS for collision avoidance services. This study addresses this gap by exploring the feasibility of integrating autonomous collision avoidance capabilities achieved within an RL framework into OOS missions.

\section{Methodology}\label{sec:method}
 
\subsection{Mission architecture}
\label{sec:3}

We consider a servicer spacecraft deployed to a parking orbit in which it constantly monitors the debris catalog. In case of potential collision risk of a target/target satellite \footnote{Target satellites are satellites that require protection from collisions with debris.}, the servicer decides autonomously, based on estimations of collision risk, to assist with CAM. It performs phasing maneuvers and docks with the target satellite, performs CAM, and returns the target satellite to its orbit. 

The current study considers a Low Earth Sun-synchronous Orbit at 750 km with an inclination of $\sim 98^\circ$ based on debris and satellite population density data \cite{Database, debris_database}. The region between 600 to 900 km is more collision-prone due to debris density, although the most populated orbit lies around 500 km.

Maneuvers for the spacecraft servicer are considered to be impulsive, assuming high thrust dynamics. Space objects are considered point objects in the simulation model. The thrust required is directly proportional to the magnitude of the velocity required to perform a certain maneuver.  Docking is assumed to be done when the relative position and velocity of the two objects are equal to zero in accordance with Lambert's solution \cite{lamberts}.

\begin{figure}
    \centering
    \includegraphics[width = 0.8\columnwidth]{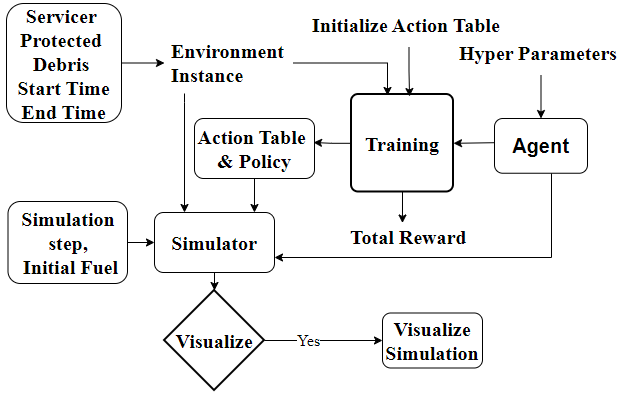}
    \caption{Simulation and training model framework.}
    \label{fig:spacenav}
\end{figure}

Fig. (\ref{fig:spacenav}) represents the training framework for the servicer to learn to perform its mission autonomously. It consists of an orbital mechanics simulator, the agent (i.e., servicer), an environment, and the agent's training algorithm. The environment refers to the simulated space environment, including the target satellites and debris. A simulator is used to generate the orbital mechanics environment and propagate the orbital elements. This is further discussed in Sec.\ref{sec:simmodel}. In RL, the agent makes decisions by acting in the environment, receiving feedback through rewards, and updating its decision policy. The agent's primary goal is to maximize the cumulative reward over time.
\subsection{Simulating the Orbital Environment}\label{sec:simmodel}
\begin{figure}
    \centering
\includegraphics[width=0.8\columnwidth]{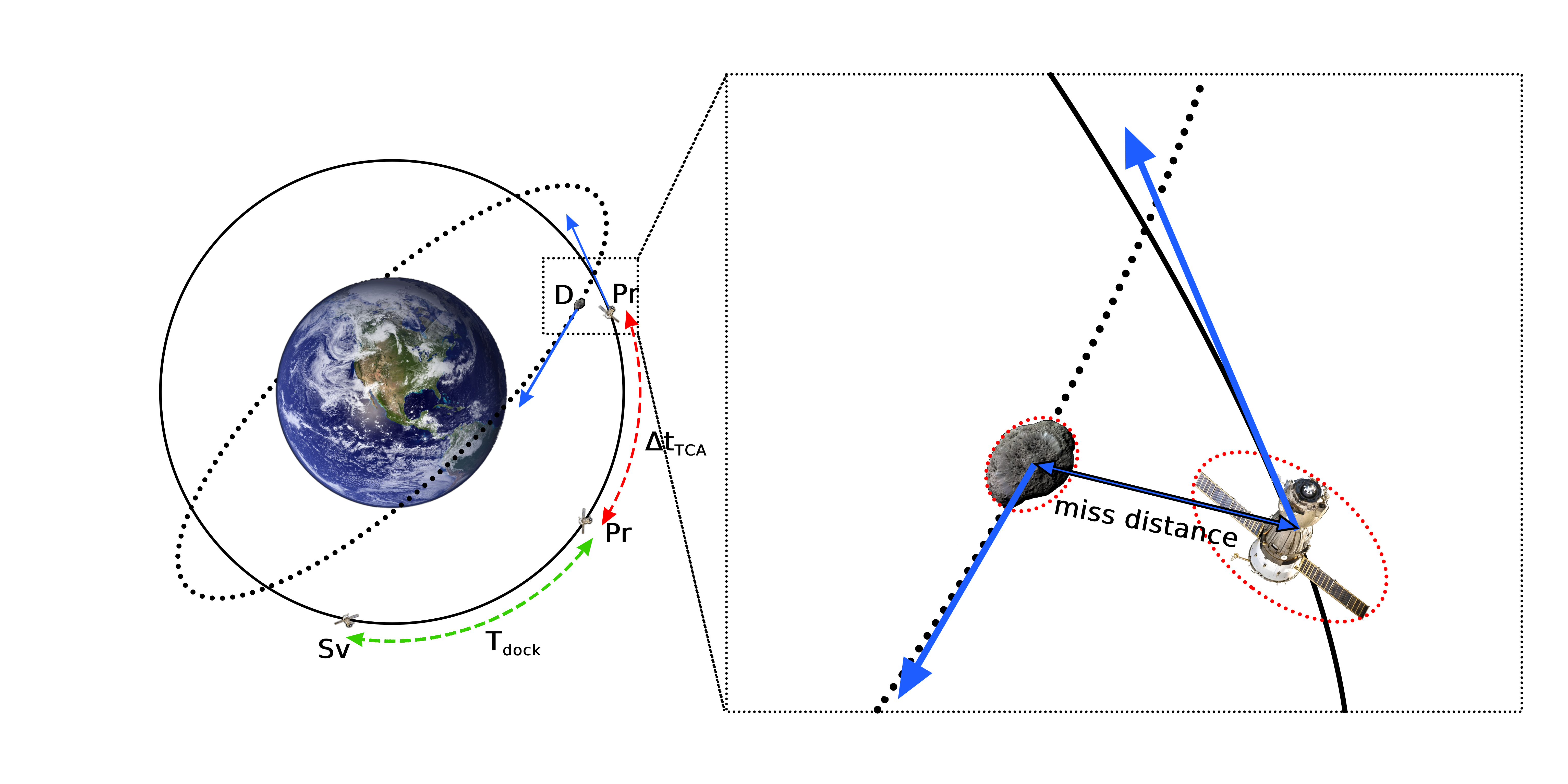}
    \caption{Schematic of a collision scenario and its relevant parameters.}
    \label{fig:scheme}
\end{figure}

A simulator based on \cite{Spacenav_lib}, which employs Keplerian orbital elements, is used in the present study to initiate the orbital elements of the satellite and debris and simulate their position over time. Keplerian ephemerides describe the motion of celestial bodies using six primary orbital elements. Using the Pykep library, the `planet.keplerian' class allows users to define celestial bodies using these Keplerian elements \cite{github_esa_pykep}. The class provides two initialization methods, one using the orbital elements directly and the other using position and velocity vectors at a specific epoch. By providing parameters like the gravitational constants of the central body and the celestial body itself, along with radii information, users can model the motion of these bodies.
 \\
The simulator can create a large number of propagation scenarios for the spacecraft and debris, providing flexibility and robustness in training. In training, collision events are created with a known time of collision, angle, and velocity ratios of debris and target satellite. \\
Fig. (\ref{fig:scheme}) represents a scenario generated by the simulator, it shows the servicer and the target satellite in the same orbit. The debris is in a generated orbit that has a potential future collision with the target satellite. The target satellite, servicer, and debris are represented by $Pr$, $Sv$, and $D$, respectively. $T_{dock}$ represents the time taken to dock with the target satellite, and $\Delta t_{TCA}$ is the time before collision, as shown in Fig. (\ref{fig:scheme}). 
\subsection{Autonomous Decision-planning model}
We model the autonomous decision-making problem as a Markov Decision Process (MDP). The MDP framework ensures that the agent makes decisions based on the current satellite state and the immediate servicing requirement, allowing for efficient and adaptive decision-making. The model elements are as follows: \\
1.) State space ($\mathcal{S}$): Represents the positions and velocities of the servicer, the target satellite, and the debris, along with the time and fuel information of the servicer.\\
\noindent2.) Action table ($\mathcal{A}$): The action table contains a sequence of maneuvers to be performed by the servicer. Each line contains the change in velocities to implement in all three directions, i.e., x,\;y,\;z, and the time (in mjd2000 format) at which the next maneuver has to be performed. It is the action table the agent refers to, and the following action is based on the state. After training, this gives the optimal maneuver matrix for the servicer to perform. \\
\noindent3.) Reward function ($\mathcal{R}$):
\noindent Extended from \cite{gremyachikh2019space}, the reward gathered by the agent is based on five elements: the collision probability, the fuel level, the trajectory deviation (for target satellite when a maneuver is performed), the relative position and velocity of the servicer and the target satellite during docking phase. Each component has a predefined threshold, and the agent receives a negative reward proportional to its deviation from the threshold \eqref{eq_reward}. 
\begin{multline}\label{eq_reward}
    R_{\text{total}} = R_{p}(P_{\text{collision}}) + R_f(dV_{\text{maneuver}}) + R_{d}(\text{deviation}) \\
   + R(\text{relative distance}) + R(\text{relative velocity})
\end{multline}

\noindent The threshold values for the deviation of the trajectory in $\{a,\;e,\;i,\;\Omega,\;\omega\}$ are based on values similar to those in \cite{gremyachikh2019space}. In particular, the threshold value considered for the probability of collision ($p_t$) is \(10^{-4}\) \cite{Nasa_CARA_Risk_Mitigation}. We employ an ELU-based function \cite{clevert2020fast} to adjust the reward function. This is done to manage the penalty for collision probabilities below a threshold $p_t$ ( $10^{-4}$) while avoiding unacceptable values ($10^{-3}$), significantly increasing the penalty for probabilities exceeding $p_t$. Furthermore, we consider values for threshold for fuel level = 500 units, deviation in \(a\) = 100m, deviation in \(e\) = 0.01, deviation in \(i\), \(\Omega\), \(\omega\) = 0.01 (in rad), docking position = 250m, docking velocity = 5m/s. The negative reward is significantly increased for each component once it crosses the threshold. All rewards are individually calculated for each component and summed up to get the total reward (Eq.~\eqref{eq_reward}). Finally, the methodology proposed in \cite{chen2017orbital}, 'Explicit Expression of $P_c$ in Terms of Conjunction Geometries,' is used to estimate the probability of collision between two objects in space. 

\subsection{RL agent: Cross-Entropy (CE) method}
\label{sec:CEmethod}
The RL model developed is solved using the CE method, which is based on a stochastic optimization approach \cite{uryasev2013stochastic}. The CE method is used to find optimal maneuvers and optimal maneuver epochs.  The first step is to choose an initial maneuver and an appropriate random distribution. The expected value $E$ of the distribution is equal to the initial maneuver parameters. The distribution is used for generating new maneuvers based on the initial one. Each maneuver is evaluated based on the reward received, and the algorithm selects those with the best rewards to adjust subsequent maneuvers. The expected value $E$ is shifted in the direction of selected maneuvers. 
Iterations are repeated until a stopping criterion is reached, which is the limit on the number of iterations, or the decision policy no longer improves. The model parameters, such as the number of iterations, number of sessions, and CE parameters, are tuned during the training. 

\section{Results}\label{sec:casestudy}
For the case study, an environment with one servicer and one target satellite is generated. The orbital parameters of the target satellite, servicer, and debris are generated to create a collision scenario, with the orbital parameters given in Table (\ref{tab:elements}). Each simulation is considered for two days, during which a potential collision with debris is generated. The start time of the experiment is ``Jan-24 21:35:59'' and the end time of the simulation is ``Jan-27 02:24:00''. The agent parameters considered for training are given as follows: number of iterations = 35, number of sessions = 30, $\sigma$ decay = 0.98, learning decay = 0.98, percentile growth = 1.005. 
\begin{table}[h!]
    \centering
    \footnotesize
    \begin{tabular}{c|c|c|c|c|c|c}
    \hline\hline
     & $a(km)$ & $e$ & $i(^\circ)$ & $ \Omega(^\circ)$ & $\omega(^\circ)$ & $\nu(^\circ)$\\\hline\hline
       $P_r$ &  7208 & 7.5e-05&324.5& 177.6& 174.3& 123\\
$S_v$ &7208 & 7.5e-05&324.5& 177.6& 174.3 & 135\\
$D$ &  7213& 7.7e-05& 13.3 & 234.3 & 330.8 & -15\\\hline\hline
    \end{tabular}
    \caption{Orbital parameters of servicer, target spacecraft, and space debris at the start of the simulation.}
    \label{tab:elements}
\end{table}

To efficiently handle the combined CAM and docking actions in training, a \textit{ hybrid time grid} approach is adopted. This approach employs a finer time grid when approaching docking to capture a close approach accurately. It switches to a coarser grid afterward for the remaining actions to balance computational efficiency while maintaining accuracy. Hence, a time step of 0.08s was assumed for docking and a coarse grid for CAM, skipping unnecessary time steps between docking and CAM maneuver.

To initialize the action table of the training agent, four maneuvers are defined. The first two maneuvers are defined for completing the docking procedures, and the last two for performing CAM and returning the target satellite to its orbit. We consider two methods to generate the initial action table: (1) random initialization, in which we start with random values for the maneuver's $\Delta V$ and timings. As the agent receives rewards from the environment, it adjusts these values to optimally perform docking and CAM if necessary. (2) Lambert solution initialization, in which we use Lambert's solution to phase the servicer with the target satellite as the initial guess for the docking maneuvers. For CAM, we start with random $\Delta V$ values and timing. This approach provides a more informed starting point compared to random initialization. At the end of the training, we compare the performance of these two initialization methods by demonstrating a simulation case study to determine their effectiveness.\\

\begin{figure}[h!]
    \centering
\includegraphics[width=0.8\columnwidth]{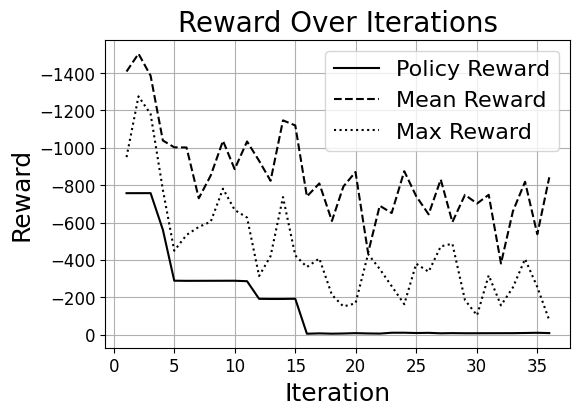}
    \caption{Reward function evolution in training - random initialization.}
    \label{fig:train_rand}
\end{figure}

\textbf{Training using random initialization:}\label{sec:randguess}
The initial guess for the action table is generated randomly for the CAM and docking maneuvers. The training process is illustrated in Fig. (\ref{fig:train_rand}), which shows the policy reward, mean, and maximum reward versus iteration count. The training plot indicates that the policy reward decreased and stabilized upon reaching the stopping criteria (the reward does not improve or the maximum number of iterations is reached). The final optimal maneuver values to be performed by the servicer for the case study are provided in Table (\ref{tab:man}).
 \begin{table}[h!]
\footnotesize
\centering
\caption{Results for the optimal maneuvers for the servicer: the case of random initialization.}
\label{tab:man}
\begin{tabular}{c|c|c|c|c}
\hline\hline
 & \textbf{$\delta V_x$(m/s)} & \textbf{$\delta V_y$(m/s)} & \textbf{$\delta V_z$(m/s)} & \textbf{t(mjd2000)} \\
\hline\hline
1 & 0.99056 & 1.42753 & -0.1519 & 6598.9000 \\
\hline
2 & 0.19245 & 0.0368 & 0.2128 & 6598.9704 \\
\hline
3 & 0.93575 & -0.2355 & -0.0597 & 6600.6005 \\
\hline
4 & -0.93575 & 0.2355 & 0.0597 & 6600.6710 \\
\hline
\end{tabular}
\end{table}
\\

\begin{figure}[h!]
     \centering
\includegraphics[width=0.8\columnwidth]{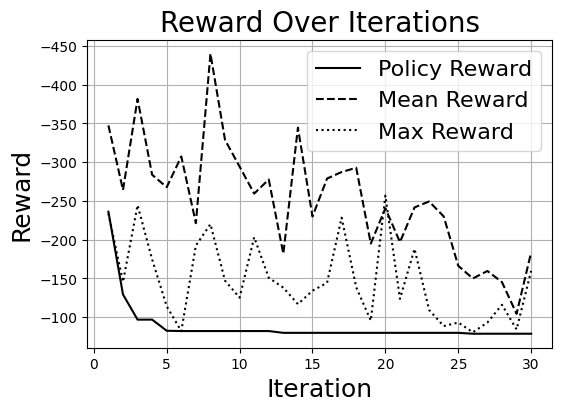}
     \caption{Reward function evolution in training - Lambert's solution initialization.}
     \label{fig:lam_train}
 \end{figure}

\begin{figure}
\centering
\begin{subfigure}{0.49\columnwidth}
  \centering
  \includegraphics[width=\linewidth]{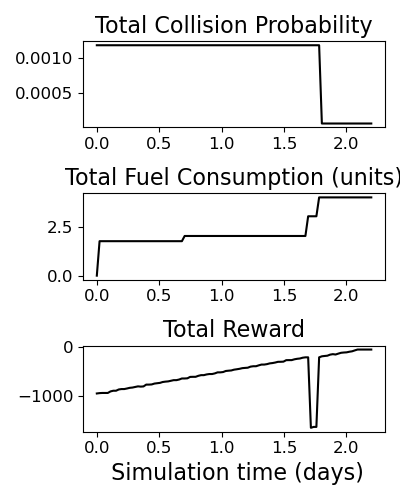}
  \caption{Random initialization}
  \label{fig:simReward}
\end{subfigure}
\begin{subfigure}{0.49\columnwidth}
  \centering
\includegraphics[width=\linewidth]{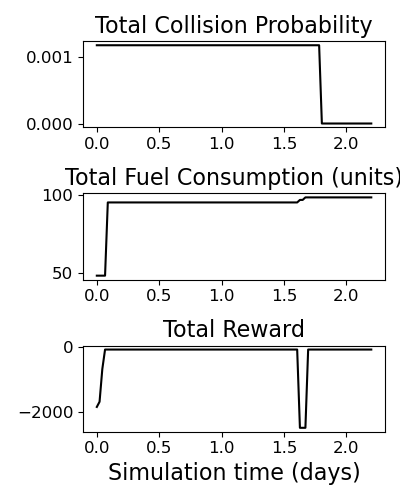}
  \caption{Lambert's initialization}
  \label{fig:simlamb}
\end{subfigure}
\caption{Results of key performance metrics.}
\label{fig:simboth}
\end{figure}

\textbf{Training using Lambert's solution initialization:}
Lambert's problem aims to find the right trajectory in space to connect two points in a specified amount of time \cite{lamberts} and is used in training the OOS vehicle in the proposed framework. A new action table is generated based on Lambert's solution initialization. It is seen that the policy reward steadily decreases with iterations and converges to the stopping criteria as seen in Fig. (\ref{fig:lam_train}). Table (\ref{tab:man_lam}) lists the optimized maneuver parameters for the simulation case study. Fig. (\ref{fig:lam_train}) further illustrates the training process, depicting the decreasing policy reward, the mean, and the maximum reward over iterations. 
\begin{table}[h!]
\footnotesize
\centering
\caption{Results for the optimal maneuvers for the servicer: the case of Lambert's solution initialization.}
\label{tab:man_lam}
\begin{tabular}{c|c|c|c|c}
\hline\hline
 & \textbf{$\delta V_x$(m/s)} & \textbf{$\delta V_y$(m/s)} & \textbf{$\delta V_z$(m/s)} & \textbf{t(mjd2000)} \\
\hline\hline
1 & $23.90$ & $32.94$ & $24.16$ & $6598.90$ \\
\hline
2 & $-18.82$ & $-35.11$ & $-25.56$ & $6598.97$ \\
\hline
3 & $0.00$ & $-0.69$ & $0.22$ & $6600.53$ \\
\hline
4 & $-0.00$ & $0.69$ & $-0.22$ & $6600.60$ \\
\hline
\end{tabular}
\end{table}
 \\
\textbf{Comparison and discussion of results:}
The model is trained in the environment described above for a large number of iterations and sessions. Training with random initialization and Lambert's solution initialization for action table and results are presented in Section \ref{sec:casestudy}. Fig. (\ref{fig:simboth}) shows the simulation results with parameter variations and variations in the reward function over time (in days) for both cases using trained models.\\
The time for the servicer to dock with the target satellite is calculated as $\approx 30$ orbital periods (between the first maneuver time and docking) as shown in Fig. (\ref{fig:simReward}). In this scenario, the docking is time-consuming due to the initial random generation of the action table for docking and CAM, resulting in velocities of approximately equal magnitude for both. \\
In practice, for CAM, maneuvers at velocities of around 1m/s are used to avoid debris and return to the original orbit \cite{gremyachikh2019space}, while phasing and docking maneuvers require an order of two or higher velocities \cite{lamberts}. \\
To resolve this issue, Training is done using Lambert's solution initialization with the results shown in Fig. (\ref{fig:simlamb}). As shown in the figure, the relative position and velocity of the docking are initially high, but it is docked in one orbit of time (a sudden decrease in the total reward function due to the successful docking). As the $\Delta$Vs are high in Lambert's initialization case, fuel consumption is high as well (Fig. \ref{fig:simboth}). The total collision probability is decreased in both scenarios to below the threshold after performing CAM, while the optimal fuel consumption is one order of magnitude higher for Lambert's solution initialization compared to random initialization. Although Lambert solution initialization facilitates faster docking and CAM execution, it comes at the expense of higher fuel consumption compared to the random initialization method during training (Fig. \ref{fig:simboth}). The sudden drop in the total reward function after 1.5 days is due to the initialization of CAM, which causes a significant deviation of the target satellite from its original orbit. The target satellite is then brought back to its orbit as the collision probability decreases and the total reward approaches zero again, as shown in Fig. \ref{fig:simboth}. \\
While the simulation assumes instantaneous docking upon the overlap of the two objects, it is important to acknowledge that a docking sequence takes several hours to initiate and complete in the real world. Consequently, decision-making within the simulation is guided by the need for proactive measures: upon detecting a potential collision, the servicer begins phasing maneuvers, aiming to dock with the target satellite. Hence, results from the model trained using Lambert's solution initialization are preferred. 
 \section{Discussion}\label{sec:conc}
This paper provides a preliminary analysis of training and using an autonomous on-orbit servicer for assisting with collision avoidance maneuvers (CAM). The case study demonstrates the ability of the servicer to autonomously make the decisions to dock and provide CAM optimally. The study simulates a scenario where there are two days before a predicted collision, focusing on the most likely orbital region for such events. This setup is close to a realistic situation where timely decision-making is crucial. Additionally, the model is trained to facilitate quick rendezvous between the servicer and the target satellite, accounting for the extra time needed in real-life scenarios to complete docking procedures. The extension of the proposed framework to cases of one servicer to multiple satellites and considering several horizons for the collision impact are also examined and will be presented in future studies.

\section*{Acknowledgment}

This work was supported by the \emph{Agence Nationale de la Recherche (ANR)}, France, under the project \textbf{ANR-23-CE10-0006} (Resilient and Sustainable Planning and Management of Future Space Industry Infrastructure with On-Orbit Servicing).
 
{\footnotesize
\bibliography{library.bib}}

\end{document}